\documentclass[a4paper,fleqn]{cas-dc}

\usepackage[authoryear,longnamesfirst]{natbib}
\usepackage{times}
\usepackage{soul}
\usepackage{url}
\usepackage[utf8]{inputenc}
\usepackage[small]{caption}
\usepackage{graphicx}
\usepackage{amsmath}
\usepackage{amsthm}
\usepackage{booktabs}
\usepackage{algorithm}
\usepackage{algorithmicx}
\usepackage{algpseudocode}
\usepackage[switch]{lineno}

\usepackage{float}
\usepackage{subfig}
\usepackage{amssymb}
\usepackage{bm}
\usepackage{indentfirst}
\usepackage{color}
\usepackage{epsfig}
\usepackage{mathtools}

\usepackage{lineno}
\usepackage{bbm}
\usepackage[capitalize]{cleveref}

\usepackage{multirow}
\def\tsc#1{\csdef{#1}{\textsc{\lowercase{#1}}\xspace}}
\tsc{WGM}
\tsc{QE}
\tsc{EP}
\tsc{PMS}
\tsc{BEC}
\tsc{DE}

\begin{document}
\let\WriteBookmarks\relax
\def\floatpagepagefraction{1}
\def\textpagefraction{.001}

\shorttitle{}

\shortauthors{Yuting Hong et~al.}

\title [mode = title]{A Multi-View Consistency Framework with Semi-Supervised Domain Adaptation
}                      



%
\author[1]{Yuting Hong}[orcid=0000-0002-1866-6746]



\ead{2211100255@nbu.edu,cn}

\credit{Conceptualization, Methodology, Software, Writing - Original Draft}

\author[1]{Li Dong}
\ead{dongli@nbu.edu.cn}
\credit{Conceptualization, Funding Acquisition, Writing - Review \& Editing}

\author[2]{Xiaojie Qiu}
\ead{xiaojie.qiu@cowain.com.cn}
\credit{Software, Writing - Review \& Editing}

\author[1]{Hui Xiao}
\ead{2011082337@nbu.edu.cn}
\credit{Investigation}

\author[1]{Baochen Yao}
\ead{2201100044@nbu.edu.cn}
\credit{Methodology, Software}

\author[3,1]{Siming Zheng}
\ead{fyyzhengsiming@nbu.edu.cn}
\credit{Data Curation, Methodology}

\author[1]{Chengbin Peng}
\cormark[1]
\ead{pengchengbin@nbu.edu.cn}
\credit{Conceptualization, Investigation, Methodology, Project Administration, Software, Writing - Original Draft, Funding acquisition, Resources, Writing - Review \& Editing, Supervision}

\cortext[cor1]{Corresponding author}

\affiliation[1]{organization={Faculty of Electrical Engineering and Computer Science, Ningbo University},
	city={Ningbo},
	postcode={315200}, 
	country={China}}

\affiliation[2]{organization={Zhejiang Cowain Automation Technology Co., Ltd.},
	city={Ningbo},
	postcode={315200}, 
	country={China}}


\affiliation[3]{organization={First Affiliated Hospital, Ningbo University},
	city={Ningbo},
	postcode={315200}, 
	country={China}}

%

\begin{abstract}
Semi-Supervised Domain Adaptation (SSDA) leverages knowledge from a fully labeled source domain to classify data in a partially labeled target domain. Due to the limited number of labeled samples in the target domain, there can be intrinsic similarity of classes in the feature space, which may result in biased predictions, even when the model is trained on a balanced dataset. To overcome this limitation, we introduce a multi-view consistency framework, which includes two views for training strongly augmented data. One is a debiasing strategy for correcting class-wise prediction probabilities according to the prediction performance of the model. The other involves leveraging pseudo-negative labels derived from the model predictions. Furthermore, we introduce a cross-domain affinity learning aimed at aligning features of the same class across different domains, thereby enhancing overall performance. Experimental results demonstrate that our method outperforms the competing methods on two standard domain adaptation datasets, DomainNet and Office-Home. Combining unsupervised domain adaptation and semi-supervised learning offers indispensable contributions to the industrial sector by enhancing model adaptability, reducing annotation costs, and improving performance.
\end{abstract}


%
%
%
%


\maketitle

\section{Introduction}

Deep convolutional networks \citep{simonyan2014very, krizhevsky2012imagenet, szegedy2017inception} have shown impressive performance in various computer vision tasks, e.g., image classification \citep{wei2021center,xu2023semi}, semantic segmentation \citep{xiao2022semi,yin2023semi} and object detection \citep{baek2024two}. These models heavily rely on a large amount of labeled data, while manual data labeling is time-consuming and labor-intensive. Thus, it is urgent to leverage unlabeled data or more easily accessible data to improve model performance. This has led to an increased interest in semi-supervised learning \citep{yang2023shrinking,liu2023wvdnet,lim2023cawm} and domain adaptation \citep{venkateswara2017deep,saito2019semi,singh2021clda,hatefi2024distribution}. 

\begin{figure}[htbp]
	\begin{center}
		\captionsetup[subfloat]{labelsep=none,format=plain,labelformat=empty}
		\subfloat{\includegraphics[width=1\linewidth]{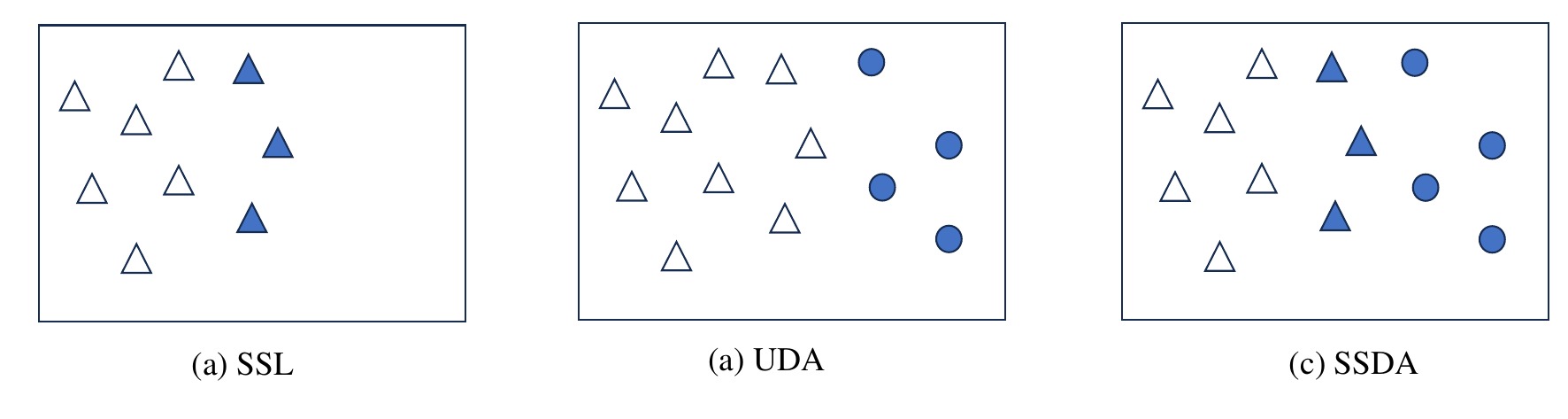}} \\
	\end{center}
	\vspace{-15pt}
	
	\caption{\textbf{An Illustration of Dataset Composition in SSL, UDA, and SSDA.} Circles represent data in the source domain, and triangles represent those in the target domain, which follow different data distributions. Data items in blue are annotated, while others are not. (a) Data in SSL are from a single domain, namely, the target domain only, which is partially annotated; (b) Data in UDA are composed of fully annotated source domain data and unlabeled target domain data. (c) Data in SSDA are composed of fully annotated source domain data and partially annotated target domain data.
}
	\label{ssda}
	\vspace{-12pt}
	
\end{figure}
Semi-Supervised Learning (SSL) is proposed to leverage an abundance of unlabeled data to enhance the model performance when labeled data is limited \citep{zhu2022introduction}, as shown in Figure \ref{ssda}(a). Consistency regularization \citep{french2019semi} and pseudo labeling \citep{hu2021simple} have shown a significant ability for leveraging unlabeled data, and thus, they are widely used in SSL frameworks. Domain Adaptation (DA) is a critical aspect of general machine learning, addressing scenarios where training and test data originate from two related but distinct domains: the source domain and the target domain \citep{venkateswara2017deep,saito2019semi,singh2021clda}. Numerous studies have extensively explored Unsupervised Domain Adaptation (UDA) shown in Figure \ref{ssda}(b) \citep{deng2019cluster}, in which labels in the target domain are inaccessible. This exploration encompasses both theoretical research \citep{ben2010theory,redko2017theoretical,zhao2019learning} and algorithmic developments \citep{ganin2016domain,kang2019contrastive,zhao2021reducing}. These approaches can utilize unlabeled data and push the boundaries of model performance.

However, SSL assumes that training and test data come from the same distribution, but in practice, this assumption often does not hold, resulting in poor model performance in new domains. In UDA, in the absence of target labeled data, the adaptation process may be unstable because there is no direct supervision signal to guide the model.

Therefore, Semi-Supervised Domain Adaptation (SSDA) has garnered increasing research attention. SSDA can effectively utilize the advantages of SSL and UDA and overcome their respective limitations by utilizing source domain labels and a limited number of target domain labels, as shown in Figure \ref{ssda}(c). These characteristics make SSDA more practical and realistic for various industrial applications. Recently, several SSL approaches \citep{grandvalet2004semi,sohn2020fixmatch} have been applied for SSDA to regularize the unlabeled data. These classic source-oriented strategies have prevailed for a long time. However, these algorithms typically require the target data to closely match some semantically similar source data in the feature space.

\begin{figure}[htbp]
	\begin{center}
		\captionsetup[subfloat]{labelsep=none,format=plain,labelformat=empty}
		\subfloat{\includegraphics[width=1\linewidth]{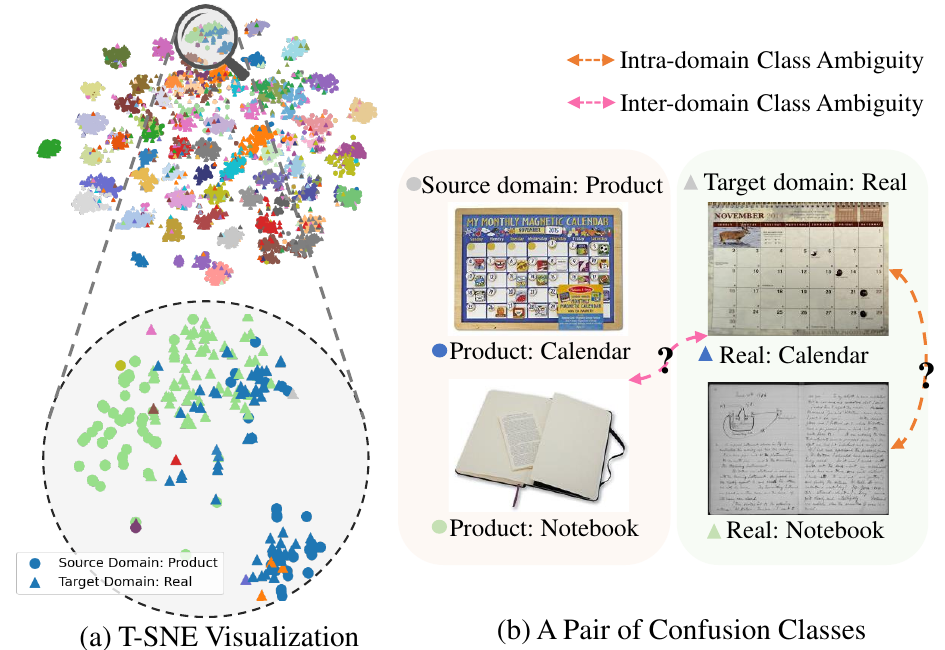}} \\
	\end{center}
	\vspace{-15pt}
	
	\caption{\textbf{T-SNE Visualization of Features.} (a) The class feature distributions of the test data after training the Baseline (S+T) on Office-Home P $\rightarrow$R with 3-shot. Circles represent the sample features in the source domain (Product), and triangles represent the sample features in the target domain (Real), with different colors indicating different classes. (b) For example, the blue triangles represent the Calendar in the target domain, which is confused not only with the Notebook in the target domain, represented by the green triangles but also with the Notebook in the source domain, represented by the green circles.
	}
	\label{confusion}
	\vspace{-12pt}
	
\end{figure}

Limited labeled data can result in the intrinsic similarity of classes, and these pseudo-labels are naturally imbalanced due to intrinsic class similarity, and the majority classes can dominate the training process. Here, the majority classes in pseudo-labels are those with a significantly higher number of pseudo-labeled instances compared to other classes, and similarly, the minority classes are those with significantly fewer instances. Training with biased and unreliable pseudo-labels can introduce errors, which may accumulate and become more pronounced as self-training progresses, leading to the Matthew effect \citep{perc2014matthew}. To address the issue of biased training, conventional methods \citep{cao2019learning,cui2019class,wei2021crest,li2020overcoming} have consistently depended on prior knowledge, such as the number of training samples per class, to formulate ad hoc sampling and weighting strategies. Nonetheless, these strategies can not resolve the imbalance issue raised by pseudo-labels which are generated dynamically during the training process, even with balanced training data. In domain adaptation, it can be even worse due to the presence of two different domains. In Figure \ref{confusion}, the Notebook class (green) is recognized as the majority class, while the Calendar class (blue) is identified as the minority class. Due to the dominance of Notebook during the training process, many Calendar samples are misclassified as Notebook, leading to ambiguity between the two classes.

\begin{figure*}[htbp]
	
	\begin{center}
		
		\includegraphics[width=1.0\textwidth]{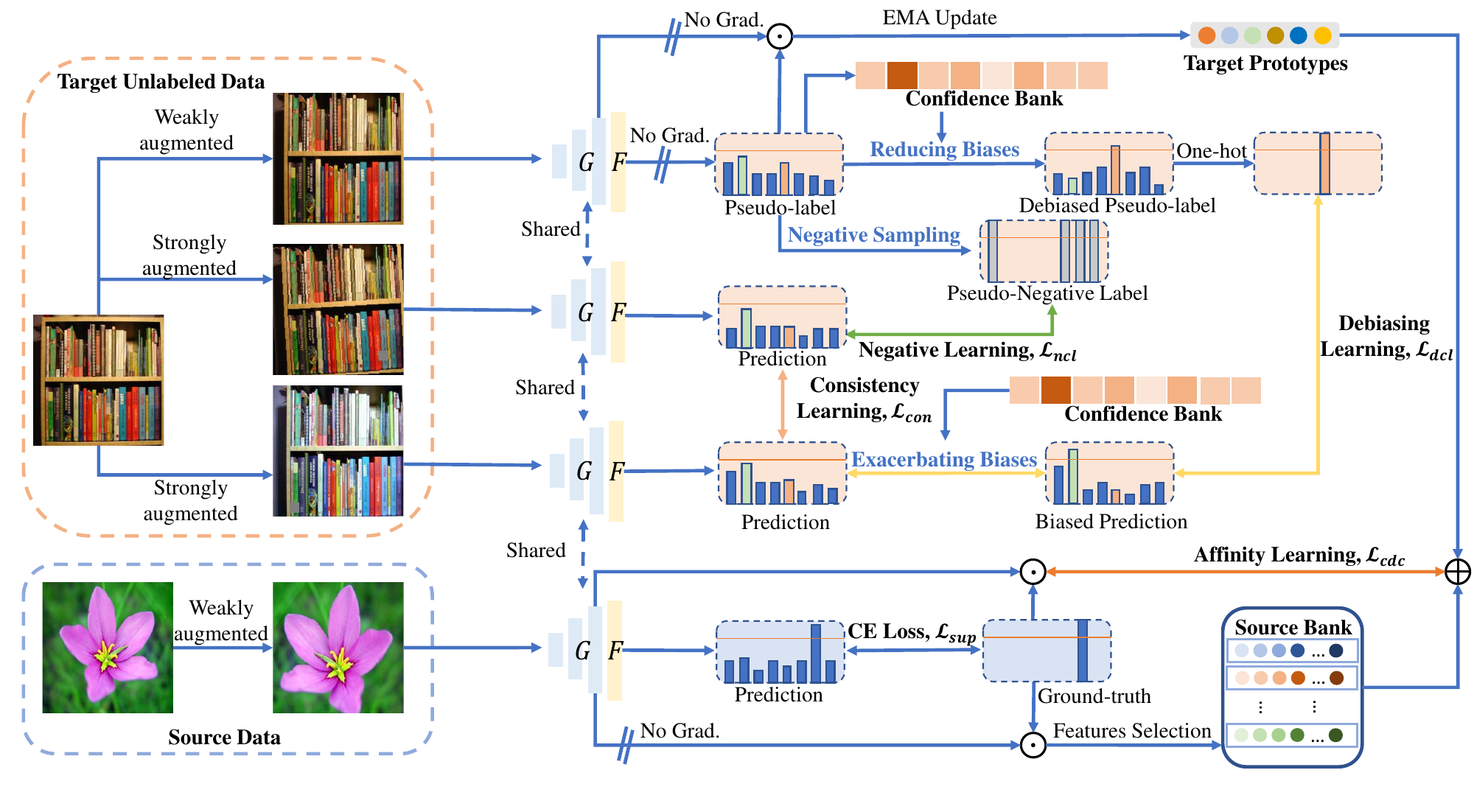}
		
	\end{center}
	\vspace{-18pt}
	
	\caption{\textbf{Overview of Our Framework.} For target unlabeled data, we first generate a weakly augmented view to feed into the model to generate the debiased pseudo-label and the pseudo-negative label, then generate two strongly augmented versions for debiasing, negative, and consistency learning. For both source data and target labeled data, we apply the traditional cross-entropy loss to the prediction after weak augmentation, then select the source domain features with the help of target prototypes for cross-domain affinity learning. Note that we omit supervised learning part for target labeled data.
	}
	\label{framework}
	\vspace{-12pt}
	
\end{figure*}

To address the aforementioned challenges, this work introduces a novel SSDA framework called \textbf{Mu}lti-\textbf{V}iew C\textbf{o}nsistency Learning (MuVo), as illustrated in Figure \ref{framework}. To the best of our knowledge, we are the first to propose consistency training with debiasing learning and negative learning to extract semantic information from two different perspectives. Thus, the model can answer both that it is probably not a Notebook and that it is more likely to be a Calendar. This approach is applied in cross-domain scenarios to mitigate biases caused by inter-class similarities, even when training on a balanced dataset.  Additionally, cross-domain affinity learning enhances the alignment of features between the source and target domains. In summary, our work has the following contributions:

\begin{itemize} 
	\item By innovatively combining debiasing learning and negative learning for consistency training, our approach allows the base model to learn from two different but well-aligned semantic views. Thus, the pseudo-label bias during training can be significantly reduced. 
	
	\item We propose a novel debiasing approach featuring adaptive probability redistribution, which dynamically adjusts the decision boundary for each class of the learning model.
	
	\item We propose a cross-domain affinity learning approach that can effectively align source domain features with target domain features, which can enhance the discrimination of class features.
	
	\item We evaluate our approach on two well-known domain adaptation datasets, namely, DomainNet and Office-Home, and our approach achieves state-of-the-art results, demonstrating its effectiveness.
\end{itemize}

\section{Related Work}
\textbf{Unsupervised Domain Adaptation} is the process of adapting a model trained on one source domain to perform well on a different target domain without using any labeled data from the target domain. In UDA, the goal is to transfer knowledge from a source domain where labeled data is available to a target domain where labeled data is scarce or unavailable so that the model can generalize effectively in the target domain. Feature alignment methods aim to minimize the global divergence \citep{chen2021enhanced,chen2019progressive,sun2016deep} between the source and target distributions. In addition, adversarial learning approaches \citep{pei2018multi,chen2019joint} have demonstrated remarkable effectiveness in attaining domain-invariant features between these domains. This method involves training the model to generate features with the purpose of misleading the domain classifier, ultimately rendering the generated features domain-agnostic. Recently, image translation techniques \citep{murez2018image}, which transform an image from the target domain to resemble an image from the source domain, have been explored in the context of UDA to address the disparities that exist between these two domains. The primary distinction between UDA and SSDA lies in the fact that SSDA allows for the utilization of a limited amount of labeled data from the target domain.

\textbf{Semi-Supervised Learning} has observed a significant advancement in recent years. Many of these methods share similar basic techniques, such as pseudo labeling \citep{lee2013pseudo, hu2021simple} or consistency regularization\citep{french2019semi, tarvainen2017mean}. Pseudo-labeling is based on a pretty natural idea that prediction obtained from model re-used as supervision, and the use of pseudo-labels is motivated by entropy minimization. Consistency regularization is based on the clustering assumption that a decision boundary typically passes through the region with low sample density. Different perturbations are added to the samples, while consistent predictions are expected. The class bias problem has been extensively studied in SSL. DST \citep{chen2022debiased} adversarially optimizes the representations to improve the quality of pseudo-labels by avoiding the worst case. BaCon \citep{feng2024bacon} directly regularizes the distribution of instances' representations in a well-designed contrastive manner. The main difference between SSL and SSDA is that SSL uses data sampled from the same distribution, while SSDA deals with data sampled from two domains with inherent domain discrepancy.

\textbf{Semi-Supervised Domain Adaptation} can be considered as a combination of UDA and SSL, where, built upon UDA, a small portion of target labeled data can be utilized. MME \citep{saito2019semi} first proposed to align the source and target distributions using adversarial training. CLDA \citep{singh2021clda} aims to narrow the intra-domain gap between the labeled and unlabeled target distributions, as well as the inter-domain gap between the source and unlabeled target distribution. DECOTA \citep{yang2021deep} decomposes SSDA into an SSL and a UDA task. The two different sub-tasks produce pseudo-labels, respectively, and learn from each other via co-training. MCL \citep{yan2022multi} introduces consistency regularization at three distinct levels for SSDA and attains outstanding results. CDAC \citep{li2021cross} introduced an adversarial adaptive clustering loss to perform cross-domain cluster-wise feature alignment. SLA \citep{yu2023semi} proposes a simple framework that efficiently constructs a label adaptation model to rectify noisy source labels. However, while solving domain adaptation problems, they do not consider the inherent biases caused by class similarity across domains. Unlike previous works, we aim to optimize the source and target domains with different perspectives to solve the more severe class bias problem of the model in the cross-domain scenario.

\section{Method}

In this section, we first establish our problem formulation in Sec \ref{Preliminary}. Then, we introduce the multi-view consistency training in Sec \ref{Pseudo-label Debiased Learning}, Sec \ref{Pseudo-label Negative Learning}, and Sec \ref{Multi-View Consistency Learning}. Finally, we describe how to use the cross-domain affinity learning in Sec \ref{Cross-Domain Cluster Learning}.

\subsection{Preliminary}
\label{Preliminary}
\textbf{Problem definition.}  In the SSDA problem, we define a fully labeled dataset  $\mathcal{D}_s=\{(x_{s},y_{s})_i\}_{i=1}^{N_s}$ from a source domain, and a partially labeled dataset from a target domain which contains a few labeled data $\mathcal{D}_{tl}=\{(x_{tl},y_{tl})_i\}_{i=1}^{N_{tl}}$ and lots of unlabeled data $\mathcal{D}_{tu}=\{(x_{tu})_i\}_{i=1}^{N_{tu}}$. The two domains share the same label space but have different data distributions. Note that the number of target labeled samples is far less than target unlabeled samples,  namely,  ${N_{tl}}\ll{N_{tu}}$. In our approach, source, target labeled, and unlabeled data are sampled equally in each training process. Our goal is to learn a task-specific classifier using $\mathcal{D}_s$, $\mathcal{D}_{tl}$ and $\mathcal{D}_{tu}$ to predict labels on test data from target domain accurately.

The SSDA model typically comprises two fundamental components: a feature extractor denoted as $\mathcal{G}$ and a classifier referred to as $\mathcal{F}$.
We first generate three versions for each target unlabeled sample $x_{tu} \in \mathcal{D}_{tu}$, denoted as $x_{tu}^{w}$, $x_{tu}^{s_1}$ and $x_{tu}^{s_2}$, through one weak and two strong augmentations, respectively. The three versions are then fed into $\mathcal{G}$ to obtain their features $f_{tu}^{w},f_{tu}^{s_1},f_{tu}^{s_2} \in \mathbb{R}^d$, and then passed through $\mathcal{F}$ to get the logit predictions $z_{tu}^{w},z_{tu}^{s_1},z_{tu}^{s_2} \in \mathbb{R}^C$ which corresponds to the outputs of the pre-softmax layer before the final predictions. By softmax function $\sigma (\cdot)$, we can get the probabilistic predictions  $p_{tu}^{w},p_{tu}^{s_1},p_{tu}^{s_2} \in \mathbb{R}^C$. This process can be formulated as follows.
\begin{align}
	f_{tu}^{\left \langle \cdot \right \rangle}=&\mathcal{G}(x_{tu}^{\left \langle \cdot \right \rangle }), \\
	z_{tu}^{\left \langle \cdot \right \rangle}=&\mathcal{F}(f_{tu}^{\left \langle \cdot \right \rangle}), \\
	p_{tu}^{\left \langle \cdot \right \rangle }=&\sigma (z_{tu}^{\left \langle \cdot \right \rangle }), \label{pred}
\end{align}
for convenience, $\left \langle \cdot \right \rangle$ can be one of $w$, $s_1$ and $s_2$.

For labeled data from the source and the target domain, each sample, represented by $x_s$ or $s_{tl}$ is fed into the same feature extractor $\mathcal{G}$ to generate features $f_s, f_{tl} \in \mathbb{R}^d$, and then fed into the classifier $\mathcal{F}$ and softmax function $\sigma (\cdot)$ to obtain the predictions  $p_s, p_{tl} \in \mathbb{R}^C$. For each sample, we minimize a standard cross-entropy (CE) classification loss as follows:
\begin{align}
\label{suploss}
	\mathcal{L}_{sup}=- (y_s \log (p_s) + y_{tl} \log (p_{tl})).
\end{align}

\subsection{Debiasing Learning}
\label{Pseudo-label Debiased Learning}
\begin{figure}[htbp]
	\begin{center}
		\captionsetup[subfloat]{labelsep=none,format=plain,labelformat=empty}
		\subfloat{\includegraphics[width=1\linewidth]{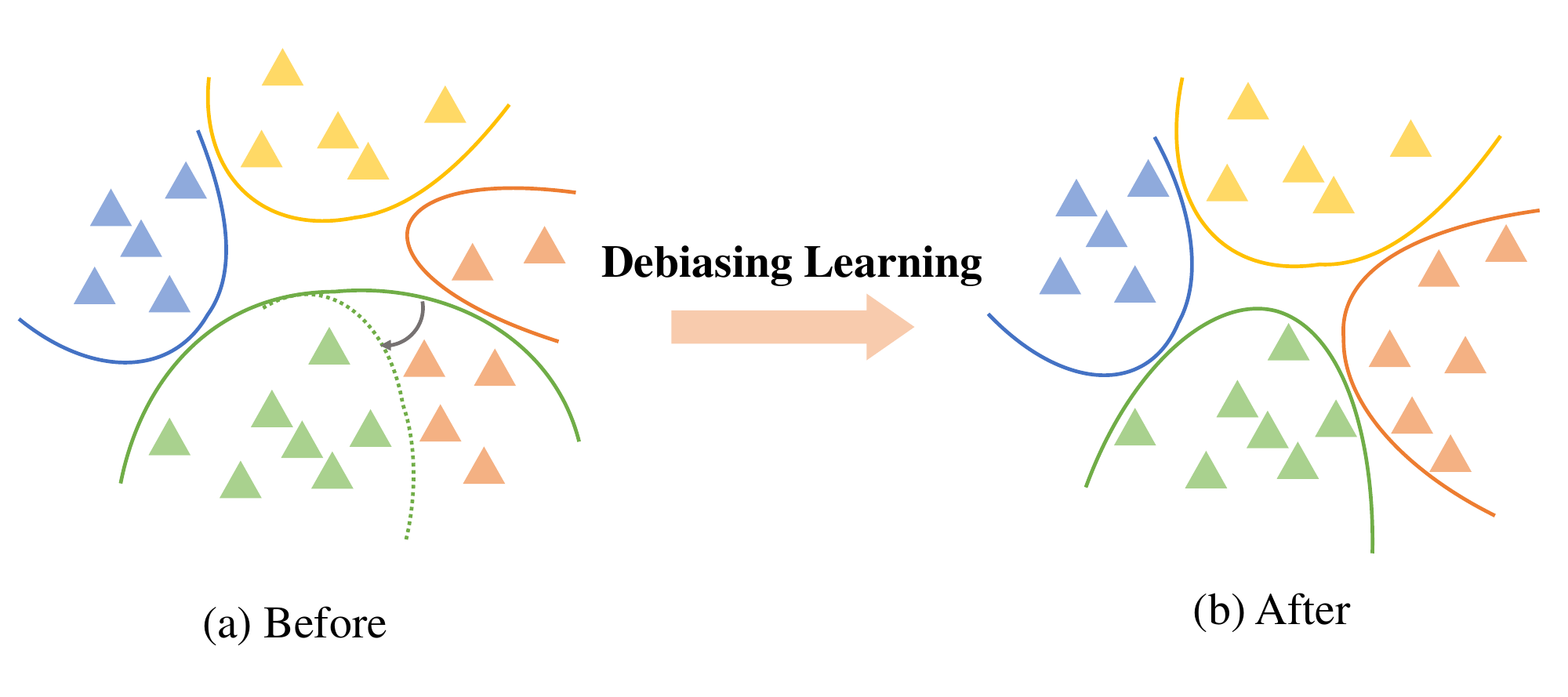}} \\
	\end{center}
	\vspace{-15pt}
	
	\caption{\textbf{An Illustration of Debiasing Learning.} Triangles indicate individual samples, with each color representing a distinct class label. Each line illustrates the decision boundary for each class. By debiasing learning, the decision boundaries for the majority classes identified by the model become smaller, and those for minority classes become larger. 
	}
	\label{debiaseddemo}
	\vspace{-12pt}
	
\end{figure}

Our approach aims at dynamically alleviating the influence of biased pseudo-labels without leveraging prior knowledge on marginal class distribution, preventing ambiguity classes from being dominated by overconfidence classes. Once ambiguous samples are assigned with wrong pseudo-labels, it is challenging to be self-corrected. On the contrary, it may even mislead the model and further amplify such biases to produce more wrong predictions. Thus, the model can get trapped in such biases without intervention. We propose an effective debiasing learning with pseudo-labels by measuring the confidence of each class on the fly without any prior knowledge about the distribution of actual classification margins.

\textbf{Confidence Bank.} We construct a global class-wise confidence bank $\Theta$ for each class and use the Exponential Moving Average (EMA) to update it at each training step:
\begin{align}
	\Theta_c\Leftarrow \lambda \Theta_c + (1-\lambda )\Theta^t_c, c\in \{0,...,C-1\},
\end{align}
where $\Theta^t_c \in (0,1)$ denotes the average confidence for Class $c$ at $t$-th iteration. Parameter $\lambda \in [0,1)$ is the momentum coefficient set to 0.999 by default. It is worth noting that we exclusively utilize samples from the target unlabeled data when computing class-wise confidence. The confidence bank can indirectly reflect the biases inherent in the model during the training process. A meager confidence value suggests subpar training performance for the given class, while a very high value implies potential overconfidence when assigning the corresponding class labels.

\textbf{Adaptive Probability Redistribution.} 
As shown in Figure \ref{debiaseddemo}, to increase the model performance on underperforming classes, we want the model to make relatively conservative predictions for the majority classes and relatively aggressive predictions for the minority classes. Motivated by this, we aim to reduce biases in weakly augmented data to generate debiased pseudo-labels, thereby better catering to minority classes. Simultaneously,  we exacerbate biases in strongly augmented data to enhance potential erroneous predictions. The redistribution gives the model a larger penalty for overconfident majority classes and further promotes the learning of underperforming classes. Consequently, the decision boundaries of the model for the majority classes become smaller, and the decision boundaries for the minority classes become larger.

First, for a debiased pseudo-label, we incorporate counterfactual reasoning \citep{pearl2009causality} to increase the confidence of the underperforming classes while decreasing the confidences of the overconfident classes. We modify the Eq.\ref{pred}, and the debiased pseudo-label is subsequently obtained as:
\begin{align}
	\tilde{p}_{tu}^{w}=\sigma(z_{tu}^{w}-\varphi \log \Theta),\label{ztuw} \\
	\tilde{y}_{tu}^{dbs}=One-Hot(\tilde{p}_{tu}^{w}),\label{onehot}
\end{align}
where the logarithmic function is to align the magnitude of $\Theta$ with that of logit, and $\varphi$ represents the debiasing factor, which governs the intensity of the indirect effect. 

Second, we adjust the prediction of the strongly augmented sample in a reversed direction, that is, increasing the confidences of the overconfident classes and decreasing the confidences of the underperforming classes. We modify the Eq.\ref{pred} and get the biased prediction as follows:
\begin{align}
	\tilde{p}_{tu}^{s_1}=\sigma(z_{tu}^{s_1} + \varphi \log \Theta).
\end{align}
When encountering an ambiguous sample, redistribution is crucial. Without proper intervention, the model may become biased towards majority classes, overlooking important patterns and nuances in minority classes. The optimization process helps the model to recognize and learn from the minority classes effectively. In addition, redistribution will have a minimal impact on those predictions that are absolutely confident. Similar to previous work \citep{sohn2020fixmatch}, we apply the cross-entropy loss to the biased prediction using the debiased pseudo-label as supervision:
\begin{align}
	\label{unsupervised}
	\mathcal{L}_{dcl}=-\mathbb{1} [\max(\tilde{p}^{w}_{tu})\geq \phi] \tilde{y}_{tu}^{dbs} \log (\tilde{p}^{s_1}_{tu}),
\end{align}
where $\mathbb{1}[\max(\tilde{p}^{w}_{tu})\geq \phi]$ means that we only select those with a confidence greater than $\phi$ as pseudo-labels for this operation.

\subsection{Negative Learning}
\label{Pseudo-label Negative Learning}

\begin{figure}[htbp]
	\begin{center}
		\captionsetup[subfloat]{labelsep=none,format=plain,labelformat=empty}
		\subfloat{\includegraphics[width=1\linewidth]{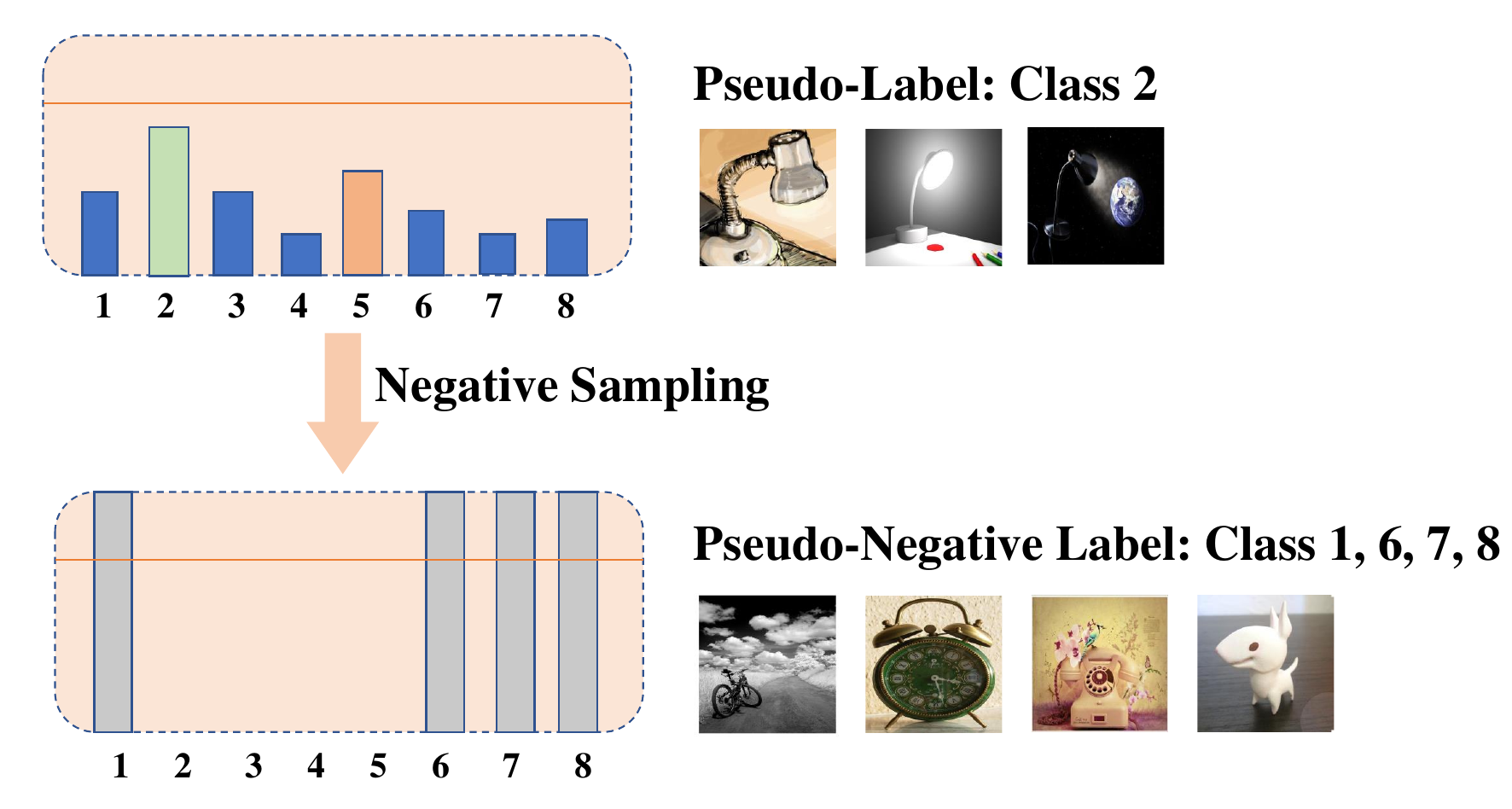}} \\
	\end{center}
	\vspace{-15pt}
	
	\caption{\textbf{An Illustration of Negative Learning.} For an weakly augmented target unlabeled data, the model makes a prediction  to generate a pseudo label (Class 2 in this example) and then  randomly selects $m$ pseudo-negative labels(Class 1, 6, 7, 8 in this example). 
	}
	\label{negativedemo}
	\vspace{-12pt}
	
\end{figure}

Negative learning methods can train convolutional neural networks with noisy data \citep{kim2019nlnl}, providing pseudo-negative labels feedback for input samples to indicate classes that do not belong. In this work, we generate pseudo-negative labels on all pseudo-labels, which differs from traditional approaches \citep{xu2023semi} that only use low-confidence pseudo-labels, and thus more semantic information can be utilized.

As shown in Figure \ref{negativedemo}, For the prediction $p_{tu}^w$ of a weakly augmented sample, we select a few pseudo-labels as representative pseudo-negative labels randomly selected from all the candidates apart from its prediction with equal probability.
\begin{align}
\label{neg1}
	\tilde{y}_{tu}^{neg} \in \psi(p_{tu}^w, m),
\end{align}
where $m$ is the number of pseudo-negative labels to select, and function $\psi$ is defined as follows:
\begin{align}
\label{neg2}
	\psi (p, m) = \{ v|v \in {\left \{ 0, 1 \right \}}^C, \sum_{i}^C v_i = m \ \text{and} \ v_{\arg\max p }\neq 1 \}.
\end{align}
By teaching with pseudo-negative labels only, we obtain semantic information different from the vanilla cross-entropy loss and eliminate the model bias in another aspect. The loss function can be written as follows:
\begin{align}
	\label{ncl}
	\mathcal{L}_{ncl}=- \tilde{y}_{tu}^{neg} \log (1 - p^{s_2}_{tu}).
\end{align}

\subsection{Multi-View Consistency Learning}
\label{Multi-View Consistency Learning}
By supervising differently augmented samples with two kinds of pseudo-labels respectively, the model can acquire distinct semantic information from two views, and by aligning samples from both views, the impact of erroneous pseudo-labels can be reduced. To allow both views to have consistent predictions, the loss function is defined as follows:
\begin{align}
	\label{con}
	\mathcal{L}_{con}= {(p^{s_1}_{tu} - p^{s_2}_{tu})}^2,
\end{align}
where $p^{s_1}_{tu}$ and $ p^{s_2}_{tu}$ denotes the softmax predictions for two strongly augmented versions of the same sample. 

\subsection{Cross-Domain Affinity Learning}
\label{Cross-Domain Cluster Learning}
To make the features more discriminative, we want the source domain features and target domain features of the same class to be closer to each other and the features of different classes to be further away from each other to further eliminate the intrinsic similarity of classes in cross-domain scenarios. Therefore, we propose cross-domain affinity learning, which can better align the source domain features with the target domain features and form clusters so that the model has a clearer decision boundary.

\textbf{Target Prototypes.} The target prototypes indicate the optimization direction for aligning the source domain features with the target domain features. We compute a target prototype $\mathcal{P}^t_c$ for Class $c$ by averaging target domain features $f_{tu}$ within the corresponding class at each training step $t$, and continuously update the global prototype $\mathcal{P}_c$ using EMA as follows:
\begin{align}
	\mathcal{P}_c\Leftarrow \lambda \mathcal{P}_c + (1-\lambda )\mathcal{P}^t_c, c\in \{0,...,C-1\}.
\end{align}

\textbf{Source Bank.} 
To maximize the utility of labeled data in the source domain, we construct a class-wise source domain feature bank $Q$ that stores highly reliable and relevant source domain features for each class. These features are crucial for better aligning source domain features with target domain features and minimizing class ambiguity. 

Firstly,  we compute the cosine similarity $sim_c$ between a source domain feature $f_s$ and each target prototype $\mathcal{P}_c$.
\begin{align}
	sim_c = \frac{f_s \cdot \mathcal{P}_c}{\|f_s\| \cdot \|\mathcal{P}_c\|} ,c \in \{0,...,C-1\},
\end{align}

Secondly, We identify this source domain feature as a candidate feature if the class corresponding to the most similar target prototype, the class predicted by the model, and the ground-truth are all consistent. Then we put these source domain features into the source bank $Q$. The oldest features are removed when the size of $Q$ exceeds the space limit.  

\textbf{Affinity Learning.}  
To have the source domain features closely align with the prototypes of the corresponding class in the target domain and simultaneously ensure a notable distance from other target prototypes, we apply a contrastive loss function  \citep{wu2018unsupervised} for each source domain feature as follows:
\begin{align}
	\label{infonce}
	\mathcal{L}_{ctr} = -\log\frac{\exp(f_s\cdot \mathcal{P}_{y_s}/\tau )}{\exp(f_s\cdot \mathcal{P}_{y_s}/\tau )+\sum_{\mathcal{P}^-_{y_s}\in \mathcal{P}}\exp(f_i\cdot \mathcal{P}^-_{{y_s}}/\tau) },
\end{align}
where $\mathcal{P}^-_{{y_s}}$ denotes the prototypes not belonging to $y_s$ and $\tau$ is a temperature hyperparameter.

In addition, we introduce a cluster loss that for strives each source domain feature $f_s$ to establish proximity to high-quality source domain features of the corresponding class $y_s$ stored in the source bank $Q$, fostering the formation of a cohesive cluster. 
\begin{align}
\label{cluster}
\mathcal{L}_{clu} = \frac{1}{\left | Q_{y_s} \right | } \sum_{f^+\in Q_{y_s}}(1-\frac{\left \langle f_s,f^+ \right \rangle }{\left \| f_s \right \|_2\cdot \left \| f^+ \right \|_2  } ),
\end{align}
where $Q_{y_s}$ means the queue of $y_s$ in source bank $Q$. 

Finally, we get the cross-domain affinity loss as follows:
\begin{align}
\mathcal{L}_{cda} =	\mathcal{L}_{ctr} + \mathcal{L}_{clu}.
\end{align}

\subsection{Training Process}
In summary, the total loss for each iteration is as follows:
\begin{align}
\mathcal{L}_{total} = \mathcal{L}_{sup} + \mathcal{L}_{dcl} +  \mathcal{L}_{ncl} + \lambda_{con}\mathcal{L}_{con} + \lambda_{cda}\mathcal{L}_{cda},
\end{align}
where $\mathcal{L}_{sup}$ is supervised loss for source data and target labeled data, $\mathcal{L}_{dcl}$ is debiasing learning loss,  $\mathcal{L}_{ncl}$ is negative learning loss,  $\mathcal{L}_{con}$ is multi-view consistency learning loss whose weight is $\lambda_{con}$ and $\mathcal{L}_{cda}$ is cross-domain affinity learning loss whose weight is $\lambda_{cda}$. 
	
The pseudo-code of MuVo is shown in algorithm \ref{algo}.

	\begin{algorithm}[h]
	\begin{minipage}{1\linewidth}
		\caption{Pseudo-code of MuVo.}
		\label{algo}
		\textbf{Input}: Labeled source data $D_s$, Labeled target data $D_{tl}$, target unlabeled data $D_{tu}$, warmup iterations $T$ and total iterations $K$.
		
		\textbf{Output}: Optimal feature extractor $\mathcal{G}$ and classifier $\mathcal{F}$.
		\begin{algorithmic}[1]
			\For{$iteration = 1,2,...,K$}
			\State {\color{blue}\# Stage \uppercase\expandafter{\romannumeral1}: Supervised learning}
			\State Compute supervised loss $ \mathcal{L}_{sup}$ by Eq. \ref{suploss};
			\State {\color{blue}\# Stage \uppercase\expandafter{\romannumeral2}: Unsupervised learning}
			\State Generate debiased pseudo-labels $\tilde{y}_{tu}^{dbs}$ by Eq.\ref{ztuw} and Eq.\ref{onehot};
			\State Generate pesudo-negative labels $\tilde{y}_{tu}^{neg}$ by Eq.\ref{neg1} and Eq.\ref{neg2};
			\State Compute $ \mathcal{L}_{dcl}$ and  $ \mathcal{L}_{ncl}$ by Eq.\ref{unsupervised} and Eq.\ref{ncl};
			\State Compute consistency loss $ \mathcal{L}_{con}$ by Eq.\ref{con};
			\State {\color{blue}\# Stage \uppercase\expandafter{\romannumeral3}: Affinity learning}
			\If{$iteration \ge T$}
			\State {\color{teal}\# Make source domain features close to the corresponding target prototypes. }
			\State Compute contrastive loss $ \mathcal{L}_{ctr}$ by Eq.\ref{infonce};
			\State {\color{teal}\# Make source domain features form clusters around the target prototypes.}
			\State Compute cluster loss $ \mathcal{L}_{clu}$ by Eq.\ref{cluster};
			\State Compute affinity loss $\mathcal{L}_{cda}$ by summing $ \mathcal{L}_{ctr}$ and $ \mathcal{L}_{clu}$;
			\EndIf
			\State \textbf{Update} $\mathcal{G}$, $\mathcal{F}$ by SGD to optimize $\mathcal{L}_{sup} + \mathcal{L}_{dcl} +  \mathcal{L}_{ncl} + \lambda_{con}\mathcal{L}_{con} + \lambda_{cda}\mathcal{L}_{cda}$.
			\EndFor
		\end{algorithmic}
	\end{minipage}
\end{algorithm}

\section{Experiments}

In this section, we evaluate the results of the proposed approach on two public datasets, Office-Home \citep{venkateswara2017deep} and DomainNet \citep{peng2019moment}. The experiments are implemented in PyTorch on a server with an NVIDIA GeForce RTX 3090.

\subsection{Experiment Setting}
\textbf{Datasets.}
Office-Home \citep{venkateswara2017deep} is a mainstream benchmark for both UDA and SSDA. The dataset contains images from four distinct domains, each representing a different office setting: Art (A), Clipart (C), Product (P), and Real (R), with 65 categories. DomainNet \citep{peng2019moment} is a large-scale dataset that serves as a valuable resource for domain adaptation and transfer learning. The dataset features images from multiple domains, each representing distinct visual styles and settings: Real (R), Clipart (C), Painting (P), and Sketch (S), with 126 classes. Besides, they focus on seven scenarios instead of combining all pairs. Our experiments follow the settings in recent works \citep{saito2019semi,yan2022multi,yu2023semi}, with the same sampling strategy for both the training set and validation set, and we conduct both 1-shot and 3-shot settings on all datasets.

\textbf{Implementation Details.} 
In line with previous works \citep{yu2023semi, li2021cross}, our experiments primarily use a ResNet34 backbone pretrained on ImageNet \citep{deng2009imagenet}. We maintain consistency with \citep{yu2023semi} by configuring parameters such as batch size, optimizer, feature size, and learning rate accordingly to ensure a fair comparison. Specifically, we set the threshold $\phi$ in Eq. \ref{unsupervised} to 0.95 and the debiasing factor $\varphi$ to 0.2. For the pseudo-negative label quantity $m$, we use 16 for the Office-Home and 36 for DomainNet. The memory bank size $M$ is set to 64 for Office-Home and 512 for DomainNet.  Due to early-stage model uncertainty, we apply negative learning for DomainNet only after the warmup stage. Following the approach outlined in \citep{yu2023semi}, the warmup parameter $T$ is set to 2000 for Office-Home and 50000 for DomainNet. For the weight of consistency learning loss, we use a ramp-up function defined as $\lambda_{con}(t) = \nu e^{-5(1-\frac{t}{T})^2}$, following \citep{laine2016temporal}. Here, $\nu$ is a scalar coefficient set to 30.0 by default, $t$ is the current iteration, and $T$ is the warmup parameter. Additionally, the affinity loss weight $\lambda_{cda}$ is set to 1.0. After the warmup stage, we refresh the learning rate scheduler, allowing the loss to be updated with a higher learning rate. It is important to note that all hyperparameters are meticulously tuned during the validation process to ensure optimal performance.

\subsection{Comparative Results}
We compare our framework with several baselines, including S+T, DANN \citep{ganin2016domain}, ENT \citep{grandvalet2004semi}, MME \citep{saito2019semi}, APE \citep{kim2020attract}, CDAC \citep{li2021cross}, DECOTA \citep{yang2021deep}, UODAv2 \citep{qin2022semi}, S$^3$D \citep{yoon2022semi}, SLA \citep{yu2023semi}, APCA \citep{ouyang2024adaptive}. S+T is a baseline method for SSDA that uses only source data and target labeled data during training. Note that all methods are implemented on a ResNet34 backbone for a fair comparison.

\begin{table*}[h]
	\caption{Accuracy (\%) on Office-Home for 1-shot and 3-shot Semi-Supervised Domain Adaptation (ResNet34).}
	\centering
	\label{Office-Home}
	\resizebox{\linewidth}{!}{
		\begin{tabular}{lccccccccccccc}
			\toprule
			\textbf{Method} & A$\rightarrow$C & A$\rightarrow$P & A$\rightarrow$R & C$\rightarrow$A & C$\rightarrow$P & C$\rightarrow$R & P$\rightarrow$A & P$\rightarrow$C & P$\rightarrow$R & R$\rightarrow$A & R$\rightarrow$C & R$\rightarrow$P & \textbf{Mean} \\ \midrule
			\multicolumn{14}{c}{\textbf{One-shot}}                                                                                                                                                                                                                  \\ \midrule
			S+T             & 50.9            & 69.8            & 73.8            & 56.3            & 68.1            & 70.0            & 57.2            & 48.3            & 74.4            & 66.2            & 52.1            & 78.6            & 63.8          \\
			DANN \citep{ganin2016domain}            & 52.3            & 67.9            & 73.9            & 54.1            & 66.8            & 69.2            & 55.7            & 51.9            & 68.4            & 64.5            & 53.1            & 74.8            & 62.7          \\
			ENT \citep{grandvalet2004semi}             & 52.9            & 75.0            & 76.7            & 63.2            & 73.6            & 73.2            & 63.0            & 51.9            & 79.9            & 70.4            & 53.6            & 81.9            & 67.9          \\
			APE \citep{kim2020attract}           & 53.9            & 76.1            & 75.2            & 63.6            & 69.8            & 72.3            & 63.6            & 58.3            & 78.6            & 72.5            & 60.7            & 81.6            & 68.9          \\
			DECOTA  \citep{yang2021deep}        & 42.1            & 68.5            & 72.6            & 60.6            & 70.4            & 70.7            & 60.6            & 48.8            & 76.9            & 71.3            & 56.0            & 79.4            & 64.8          \\
			MME  \citep{saito2019semi}            & 59.6            & 75.5            & 77.8            & 65.7            & 74.5            & 74.8            & 64.7            & 57.4            & 79.2            & 71.2            & 61.9            & 82.8            & 70.4          \\
			CDAC \citep{li2021cross}           & 61.2            & 75.9            & 78.5            & 64.5            & 75.1            & 75.3            & 64.6            & 59.3            & 80.0            & 72.9            & 64.1            & 83.8            & 72.0          \\ 
			UODAv2 \citep{qin2022semi}        & 44.9           & 71.0           & 73.2            & 58.7          & 72.8           & 70.6           & 60.5          & 49.7            & 75.9       & 66.9          & 51.6          & 80.9           & 64.7          \\
			S$^3$D \citep{yoon2022semi}         & 59.3           & 75.3           & 77.4            & 64.6          & 76.1           & 73.6           & 64.7           & 56.8            & 79.0           & 71.0           & 63.2           & 82.3           & 70.3          \\
			SLA \citep{yu2023semi}            & 63.0            & 78.0            & 79.2            & 66.9            & 77.6            & 77.0            & 67.3            & 61.8            & 80.5            & 72.7            & 66.1            & \textbf{84.6}            & 72.9          \\ \midrule
			MuVo (Ours)  &   \textbf{63.4}              &      \textbf{79.4}           &     \textbf{80.1} &     \textbf{67.7}          &   \textbf{79.2}              &   \textbf{77.3}              &   \textbf{67.5}              &   \textbf{63.5}              &     \textbf{81.6}       &     \textbf{73.2}            &  \textbf{66.6}               &        84.5         &   \textbf{73.7}          \\ \midrule
			\multicolumn{14}{c}{\textbf{Three-shot}}                                                                                                                                                                                                                \\ \midrule
			S+T             & 54.0            & 73.1            & 74.2            & 57.6            & 72.3            & 68.3            & 63.5            & 53.8            & 73.1            & 67.8            & 55.7            & 80.8            & 66.2          \\
			DANN \citep{ganin2016domain}           & 54.7            & 68.3            & 73.8            & 55.1            & 67.5            & 67.1            & 56.6            & 51.8            & 69.2            & 65.2            & 57.3            & 75.5            & 63.5          \\
			ENT \citep{grandvalet2004semi}             & 61.3            & 79.5            & 79.1            & 64.7            & 79.1            & 76.4            & 63.9            & 60.5            & 79.9            & 70.2            & 62.6            & 85.7            & 71.9          \\
			APE \citep{kim2020attract}            & 63.9            & 81.1            & 80.2            & 66.6            & 79.9            & 76.8            & 66.1            & 65.2            & 82.0            & 73.4            & 66.4            & 86.2            & 74.0          \\
			DECOTA  \citep{yang2021deep}        & 64.0            & 81.8            & 80.5            & 68.0            & 83.2            & 79.0            & 69.9            & 68.0            & 82.1            & 74.0            & 70.4            & 87.7            & 75.7          \\
			MME \citep{saito2019semi}             & 63.6            & 79.0            & 79.7            & 67.2            & 79.3            & 76.6            & 65.5            & 64.6            & 80.1            & 71.3            & 64.6            & 85.5            & 73.1          \\
			CDAC \citep{li2021cross}           & 65.9            & 80.3            & 80.6            & 67.4            & 81.4            & 80.2            & 67.5            & 67.0            & 81.9            & 72.2            & 67.8            & 85.6            & 74.8          \\ 
			UODAv2 \citep{qin2022semi}        & 55.7          & 78.6          & 75.3        & 63.3          & 78.9          & 74.0          & 61.8        & 56.8           & 78.3       & 68.0          & 59.3      & 83.6           & 69.5         \\
			SLA \citep{yu2023semi}            & 67.3            & \textbf{82.6}            & 81.4            & 69.2            & 82.1            & 80.1            & \textbf{70.1}            & 69.3            & 82.5            & 73.9            & 70.1           & 87.1            & 76.3          \\ \midrule
			MuVo (Ours)  &       \textbf{68.8}          &       82.2           &    \textbf{81.9}     &    \textbf{69.6}   & \textbf{82.7}     &  \textbf{80.5}  &   70.0          &  \textbf{69.5}               &     \textbf{83.1}            &     \textbf{74.0}            &   \textbf{70.2} &       \textbf{87.9}      &    \textbf{76.7}           \\ \bottomrule
	\end{tabular}}
\end{table*}
\begin{table*}[h]
	\caption{Accuracy (\%) on DomainNet for 1-shot and 3-shot Semi-Supervised Domain Adaptation (ResNet34).}
	\centering
	\label{DomainNet}
	\resizebox{\linewidth}{!}{
		\begin{tabular}{lcccccccccccccccc}
			\toprule
			& \multicolumn{2}{c}{R$\rightarrow$C} & \multicolumn{2}{c}{R$\rightarrow$P} & \multicolumn{2}{c}{P$\rightarrow$C} & \multicolumn{2}{c}{C$\rightarrow$S} & \multicolumn{2}{c}{S$\rightarrow$P} & \multicolumn{2}{c}{R$\rightarrow$S} & \multicolumn{2}{c}{P$\rightarrow$R} & \multicolumn{2}{c}{\textbf{Mean}} \\
			\textbf{Method} & 1-shot           & 3-shot           & 1-shot           & 3-shot           & 1-shot           & 3-shot           & 1-shot           & 3-shot           & 1-shot           & 3-shot           & 1-shot           & 3-shot           & 1-shot           & 3-shot           & \textbf{1-shot} & \textbf{3-shot} \\ \midrule
			S+T             & 55.6             & 60.0             & 60.6             & 62.2             & 56.8             & 59.4             & 50.8             & 55.0             & 56.0             & 59.5             & 46.3             & 50.1             & 71.8             & 73.9             & 56.9            & 60.0            \\
			DANN \citep{ganin2016domain}             & 58.2             & 59.8             & 61.4             & 62.8             & 56.3             & 59.6             & 52.8             & 55.4             & 57.4             & 59.9             & 52.2             & 54.9             & 70.3             & 72.2             & 58.4            & 60.7            \\
			ENT \citep{grandvalet2004semi}            & 65.2             & 71.0             & 65.9             & 69.2             & 65.4             & 71.1             & 54.6             & 60.0             & 59.7             & 62.1             & 52.1             & 61.1             & 75.0             & 78.6             & 62.6            & 67.6            \\
			APE \citep{kim2020attract}             & 70.4             & 76.6             & 70.8             & 72.1             & 72.9             & 76.7             & 56.7             & 63.1             & 64.5             & 66.1             & 63.0             & 67.8             & 76.6             & 79.4             & 67.6            & 71.7            \\
			DECOTA \citep{yang2021deep}         & 79.1             & 80.4             & 74.9             & 75.2             & 76.9             & 78.7             & 65.1             & 68.6             & 72.0             & 72.7             & 69.7             & 71.9             & 79.6             & 81.5             & 73.9            & 75.6            \\
			MME \citep{saito2019semi}             & 70.0             & 72.2             & 67.7             & 69.7             & 69.0             & 71.7             & 56.3             & 61.8             & 64.8             & 66.8             & 61.0             & 61.9             & 76.1             & 78.5             & 66.4            & 68.9            \\
			CDAC \citep{li2021cross}          & 77.4             & 79.6             & 74.2             & 75.1             & 75.5             & 79.3             & 67.6             & 69.9             & 71.0             & 73.4             & 69.2             & 72.5             & 80.4             & 81.9             & 73.6            & 76.0            \\
			UODAv2 \citep{qin2022semi}            & 77.0           & 79.4            & 66.8             & 68.2          & 61.6         & 67.2          & 55.3             & 61.1       & 63.4           & 65.5           &52.8           & 59.4           & 73.6           & 76.6       & 62.6      & 66.5         \\ 
			S$^3$D \citep{yoon2022semi}            & 73.3             & 75.9             & 68.9             & 72.1            & 73.4             & 75.1             & 60.8             & 64.4       & 68.2           & 70.0            & 65.1             & 66.7             & 79.5           & 80.3             & 69.9        & 72.1           \\ 
			SLA \citep{yu2023semi}            & \textbf{79.8}             & \textbf{81.6}             & 75.6             & 76.0             & 77.4             & 80.3             & 68.1             & 71.3             & 71.7             & 73.5             & 71.7             & 73.5             & 80.4             & 82.5             & 75.0            & 76.9            \\  
			APCA \citep{ouyang2024adaptive}            & 77.4             & 79.4             & 74.2             & 76.8             & 77.1             & 79.9             & \textbf{70.2}           & 71.7             & \textbf{73.3}             & 73.4             & 70.1             & 71.9             & 80.2             & 81.8             & 74.6            & 76.4            \\
			\midrule
			MuVo (Ours)      &   79.5               &      80.6            &       \textbf{76.2}          &    \textbf{76.7}       &   \textbf{77.9}               &     \textbf{80.6}        &      69.4       &    \textbf{72.1}              &   72.7            &     \textbf{74.5}      &       \textbf{72.2}           &        \textbf{73.7}          &    \textbf{81.8}             &   \textbf{83.6}               &     \textbf{75.7}            &    \textbf{77.4}             \\ \bottomrule
	\end{tabular}}
\end{table*}

\textbf{Results on Office-Home Dataset.} We verified the effectiveness of our method on Office-Home. As shown in Table \ref{Office-Home}, MuVo consistently outperforms other methods on both the 1-shot and 3-shot settings and on almost all adaptation scenarios. The mean accuracy of MuVo reaches 73.7\% and 76.7\%, which outperforms the baseline by around 9.9\% and 10.5\%, under the 1-shot and 3-shot settings, respectively.

\textbf{Results on DomainNet Dataset.} We verified the effectiveness of our method on DomainNet. Table \ref{DomainNet} shows the comparative results. To our knowledge, our method outperforms existing methods with significant superiority except for only one case. Overall, we get 75.7\% and 77.4\% mean accuracy over seven adaptation scenarios under the 1-shot and 3-shot settings, respectively. The limitations in certain cases, such as A$\rightarrow$P for Office-Home and R$\rightarrow$C for DomainNet, probably arise from the substantial feature differences between these domains, making it more difficult to align features from different domains. In such a case, domain transfer often requires more sophisticated feature alignment methods.

\subsection{Analysis}
In this part, we investigate the performance gain of each proposed module through some experiments. All the following experiments are performed on Office-Home A $\rightarrow$ C, C$\rightarrow $P and P$\rightarrow $R. All experiments are implemented on a ResNet34 backbone.

\textbf{Ablation Studies.} We begin by showcasing the contributions of each proposed component to the SSDA performance on Office-Home A$\rightarrow $C, C$\rightarrow $P and P$\rightarrow $R, and under 3-shot settings as shown in Table \ref{ablation}.  When a certain loss is not used in a specific ablation experiment, we remove it from the total loss during training. In this table, each module significantly improves the performance of SSDA. These enhancements are particularly notable compared to the baseline S+T, which is trained only on source and labeled target data. For A$\rightarrow $C, the debiasing learning $\mathcal{L}_{dcl}$ independently achieves an accuracy of 66.5\%, resulting in a notable improvement of 12.5\%; this is mainly attributed to the support of unlabeled data and aggressive training on minority classes. Simultaneously, introducing pseudo-negative label learning $\mathcal{L}_{ncl}$ degrades performance slightly; this may be due to the aggravation of the consequences of selecting ground-truth as pseudo-negative labels by mistake when there are no other constraints and the quantity of pseudo-negative labels is large. Motivated by this, we added a consistency constraint to the two views to prevent one view from optimizing in the wrong direction, thereby producing positive training. Upon incorporating multi-view consistency learning $\mathcal{L}_{con}$, the accuracy further enhances 2.3\%. Finally, feature alignment plays a vital role in domain adaptation tasks, as it can improve the model's generalization ability and enhance performance. Inspired by this, the inclusion of the cross-domain affinity learning $\mathcal{L}_{cda}$ contributes to an overall improvement of 14.8\% in our final results. For C$\rightarrow $P and P$\rightarrow $R, we find that each module has a different degree of promotion. The introduction of debiasing learning $\mathcal{L}_{dcl}$ results in a performance improvement of 9.1\% and 9.3\%, respectively. Implementing negative learning $\mathcal{L}_{ncl}$ further enhances the performance by 0.5\% and 0.2\%. Additionally, consistency learning $\mathcal{L}_{con}$ leads to an incremental improvement of 0.4\% and 0.3\%. Finally, with the integration of affinity learning $\mathcal{L}_{cda}$, the performance reaches its peak, achieving 82.7\% and 83.1\%, respectively. In conclusion, MuVo shows great potential for enhancing efficiency in SSDA tasks. 

\textbf{Comparison of Different View Combinations.} We explore the pairwise comparisons of three different views: the original view ($\tilde {y}_{tu} $), the debiased view ($\tilde {y}_{tu}^{dbs} $) and the negative view ($\tilde {y}_{tu}^{neg} $) in Table \ref {views}. Row (b) and (c) tend to exhibit better performance than Row (d) and (e); that is, when using two identical views, the effect is often inferior to the original view combined with the debiased view or negative view, which may be due to the similarity of semantic information leading to suboptimal performance. In particular, row(d) exhibits the worst performance, only 65.0\%, which may be due to the fact that the combination of the two debiased views makes the model predictions more radical. The combination of debiased view and negative view yields the best performance, reaching 68.8\%, as it fully considers the minority classes and captures more semantic information.

\begin{table}[htb]
	\caption{Ablation studies of MuVo’s different components. We report the Accuracy (\%) on Office-Home A $\rightarrow$ C, C $\rightarrow$ P and P $\rightarrow$ R under the settings of 3-shot using a ResNet34 backbone.}
	\label{ablation}
	\centering
	\scalebox{0.8}{
		\begin{tabular}{cccccc|ccc}
			\toprule
			& $\mathcal{L}_{sup}$    & $\mathcal{L}_{dcl}$    & $\mathcal{L}_{ncl}$    & $\mathcal{L}_{con}$    & $\mathcal{L}_{cda}$                         & A $\rightarrow$ C & C $\rightarrow$ P & P $\rightarrow$ R \\ \midrule
			$(a)$ & $\checkmark$ &             &         &              & \multicolumn{1}{l|}{}             & 54.0 &72.3&72.9    \\
			$(b)$ & $\checkmark$ & $\checkmark$ &              &              & \multicolumn{1}{l|}{}             & 66.5  & 81.4 &82.2  \\
			$(c)$ & $\checkmark$ & $\checkmark$ & $\checkmark$ &              & \multicolumn{1}{l|}{}             & 66.2  &81.9&82.4   \\
			$(d)$ & $\checkmark$ & $\checkmark$ & $\checkmark$ & $\checkmark$ & \multicolumn{1}{l|}{}             & 68.5  & 82.3 & 82.7   \\
		\rowcolor{Gray!20}	$(e)$ & $\checkmark$ & $\checkmark$ & $\checkmark$ & $\checkmark$ & \multicolumn{1}{l|}{$\checkmark$} & \textbf{68.8}  &\textbf{82.7}&\textbf{83.1}    \\ \hline
		\end{tabular}
	}
\end{table}

\begin{table}[h]
	\caption{Comparison of combinations of different views. We report the Accuracy (\%) on Office-Home A $\rightarrow$ C under the settings of 3-shot using a ResNet34 backbone.}
	\label{views}
	\centering
	\scalebox{1}{
		\begin{tabular}{cccc|ccc|c}
			\toprule
			\multicolumn{4}{c|}{$View\ A$}  & \multicolumn{3}{c|}{$View\ B$} & \multirow{2}{*}{Accuracy(\%)} \\
			&$\tilde{y}_{tu}$& $\tilde{y}_{tu}^{dbs}$& $\tilde{y}_{tu}^{neg}$& $\tilde{y}_{tu}$& $\tilde{y}_{tu}^{dbs}$& $\tilde{y}_{tu}^{neg}$&  \\ 
			\midrule
			$(a)$&$\checkmark$  &   &    & $\checkmark$  &     &    & 67.3  \\
			$(b)$&$\checkmark$  &     &   &   & $\checkmark$    &   & 68.3  \\
			$(c)$&$\checkmark$  &   &   &  &   & $\checkmark$       & 67.6   \\
			$(d)$&& $\checkmark$   &    &     & $\checkmark$     &  & 65.0   \\
			$(e)$&&    & $\checkmark$&  &   & $\checkmark$    & 67.2    \\
			\rowcolor{Gray!20}$(f)$&& $\checkmark$   &   &    &  & $\checkmark$   & \textbf{68.8}   \\ \hline
		\end{tabular}
	}
\end{table}

\begin{figure}[htbp]
	\begin{center}
		\captionsetup[subfloat]{labelsep=none,format=plain,labelformat=empty}
		\subfloat{\includegraphics[width=1\linewidth]{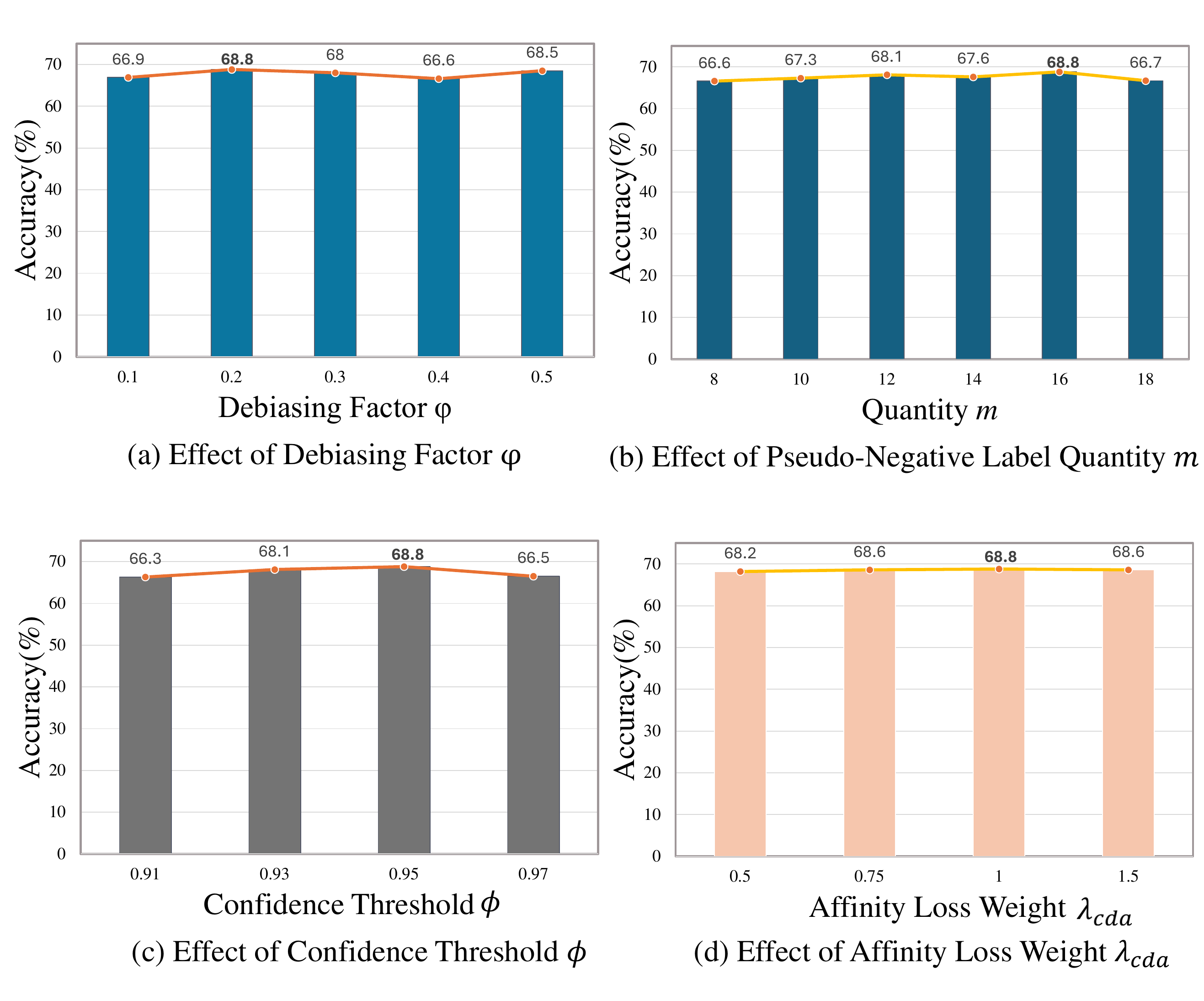}} \\
	\end{center}
	\vspace{-15pt}
	
	\caption{\textbf{Effects of Hyperparameters.} (a) Effect of debiasing factor $\varphi$. (b) Effect of pseudo-negative label quantity $m$. (c) Effect of confidence threshold $\phi$. (d) Effect of affinity loss weight $\lambda_{cda}$. All results are obtained on Office-Home A $\rightarrow$ C with 3-shot and ResNet34 backbone.
	}
	\label{zhexian}
	\vspace{-10pt}
	
\end{figure}

\textbf{Effect of Debiasing Factor $\varphi$.} As illustrated in Figure  \ref{zhexian}(a). The model is not sensitive to the debiasing factor in a specific range and shows fluctuations while the overall performance is improved. In short, introducing a pseudo-label debiasing learning demonstrates a performance improvement, particularly when opting for the optimal $\varphi$ value of 0.2. We kept 0.2 throughout all experiments for simplicity and resource-saving. However, a tradeoff exists. If the debiasing factor $\varphi$ is too large, the model may struggle to adapt effectively to the data. In contrast, an excessively small factor may inadequately address biases, ultimately impairing the model's generalization ability.

\textbf{Effect of Pseudo-Negative Label Quantity $m$.} 
As indicated in Figure \ref{zhexian}(b), the quantity $m$ is not highly sensitive to the performance of the model when all settings are consistent. However, slightly larger or smaller values can have a minor impact on the model performance. It is crucial to have an appropriate number of pseudo-negative labels for optimal results. Suppose the value of $m$ is excessively large, such as in the case of Office-Home, which approaches a quarter of the total number of categories. In that case, there is a risk of selecting ground-truth, potentially compromising the overall performance.

\textbf{Effect of Confidence Threshold $\phi$.} In SSL, balancing data utilization and pseudo-label quality is crucial when setting the pseudo-label threshold. Setting the threshold too high results in low data utilization, which weakens the model's generalization ability. Conversely, setting the threshold too low introduces more noise, which misleads the model's decision boundary. An appropriate threshold ensures that high-quality pseudo-labels are used for training, maximizing the use of unlabeled data. As demonstrated in Figure \ref{zhexian}(c), we evaluate the impact of different confidence thresholds on model performance. Our approach is not sensitive to this hyperparameter within a specific range. The results indicate that the accuracy increases with higher thresholds up to a point. Specifically, at a threshold of 0.91, the accuracy is 66.3\%. Increasing the threshold to 0.93 results in an improved accuracy of 68.1\%, and raising it to 0.95 yields the highest accuracy of 68.8\%. However, setting the threshold to 0.97 leads to a performance drop to 66.5\%. These results suggest that while increasing the confidence threshold can enhance performance, overly stringent thresholds may adversely affect the outcome.

\textbf{Effect of Affinity Loss Weight $\lambda_{cda}$.} 
As depicted in Figure \ref{zhexian}(d),  we investigate the impact of different affinity loss weights on model performance. The loss weight is not sensitive to model performance as a whole within a specific range. When the loss weight was set to 0.5, the model achieved an accuracy of 68.2\%. Setting the weight to 0.75 resulted in a slight increase in accuracy to 68.6\%, and when the weight is 1, the accuracy is 68.8\%. When the weight is 1.5, there is a slight decrease in accuracy, to be 68.6\%. These results indicate that the model is not sensitive to the parameter setting, with a small performance fluctuation when the weight varies between 0.5 and 1.5.

\begin{figure}[htbp]
	\begin{center}
		\captionsetup[subfloat]{labelsep=none,format=plain,labelformat=empty}
		\subfloat{\includegraphics[width=1\linewidth]{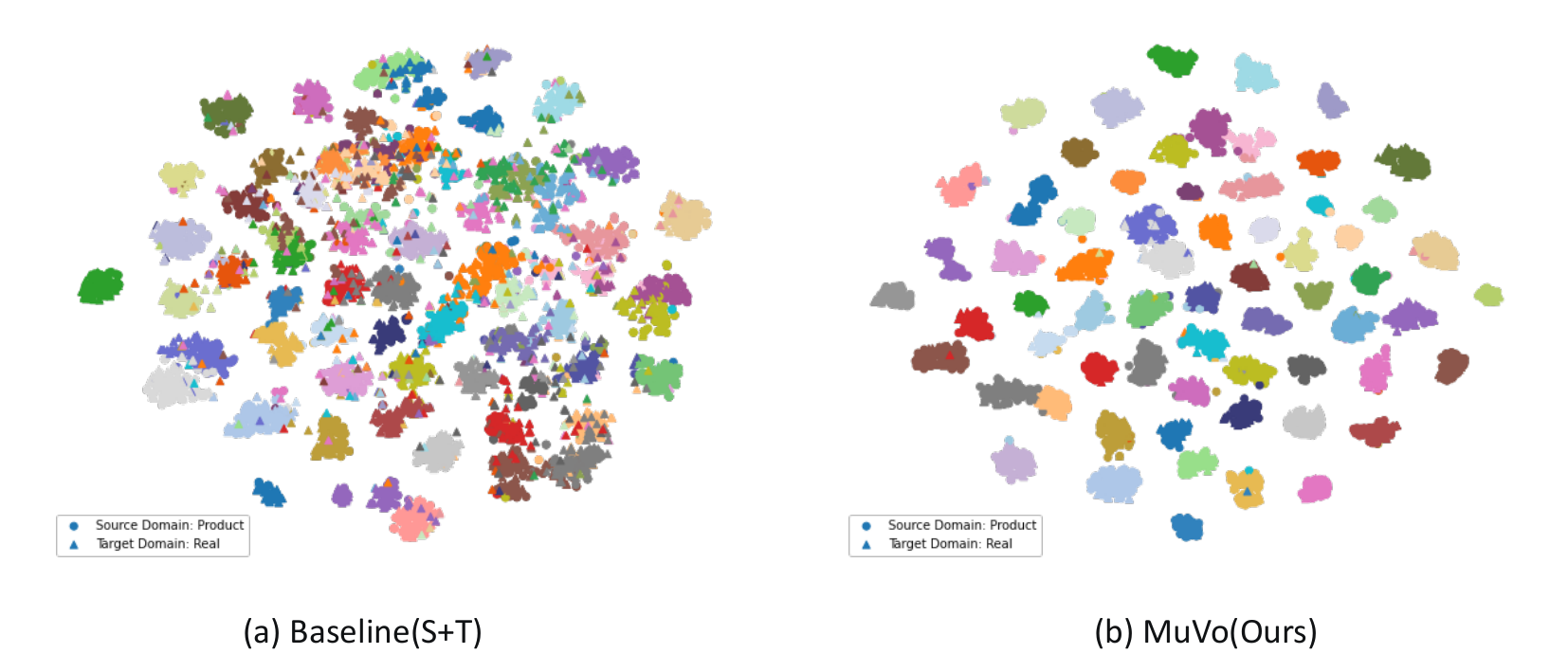}} \\
	\end{center}
	\vspace{-15pt}
	
	\caption{\textbf{Comparison of T-SNE Visualization.} (a) The class feature distributions after training with the Baseline (S+T). (b) The class feature distributions after training with MuVo (Ours). Circles represent sample features in the source domain (Product), and triangles represent sample features in the target domain (Real), with different colors indicating different classes. The features are derived from the test data for both domains. The optimization achieved by our method is evident. All results are obtained on the Office-Home P $\rightarrow$ R with a 3-shot setting and a ResNet34 backbone.
	}
	\label{tsne}
	\vspace{-12pt}
	
\end{figure}

\textbf{Comparison of T-SNE Visualization.} We compare test data feature distributions of two domains produced by the baseline and MuVo after T-SNE visualization. In Figure \ref{tsne}, the source domain features and target domain features have been well-aligned and formed clusters. The model's decision boundary becomes clearer, leading to more accurate predictions. Consequently, The intra-domain ambiguity and inter-domain ambiguity between different classes are effectively eliminated. 

\section{Conclusion}
In this paper, we introduce a novel framework for SSDA that combines debiasing learning and negative learning for consistency training and also introduce cross-domain affinity learning. In real-world applications, models often exhibit bias, even with balanced datasets, let alone imbalanced ones. We apply our framework in the cross-domain scenario to solve the class bias problem caused by the intrinsic similarity of classes. This area has yet to be extensively studied before. This work offers valuable insights into the SSDA field, paving the way for further research and practical applications in real-world scenarios. Our experimental results on the DomainNet and Office-Home datasets demonstrate the effectiveness of our method, outperforming existing techniques. Additionally, our approach is designed for seamless integration into various SSDA industrial tasks. Nevertheless, as with other SSL approaches, the training process of our method is more time-consuming than that of fully supervised methods. Also, when the feature difference between different domains is relatively significant, it makes the cross-domain features more difficult to align, and thus, in such scenarios, more sophisticated feature alignment methods in domain transfer can be explored in the future.

\printcredits

\bibliographystyle{model1-num-names}

\bibliography{cas-refs}


\end{document}